%% file: acl.tex
\pdfoutput=1
\documentclass[11pt]{article}
\usepackage{acl}
\usepackage{times}
\usepackage{latexsym}
\usepackage[T1]{fontenc}
\usepackage[utf8]{inputenc}

\usepackage{microtype}
\usepackage{booktabs}
\usepackage{multirow}
\usepackage{multicol}
\usepackage{subfigure}
\usepackage{graphicx}
\usepackage[linesnumbered,ruled]{algorithm2e}
\usepackage{mathtools,amssymb,mathrsfs}
\usepackage{microtype}
\usepackage{enumitem}
\usepackage{pdfpages}
\usepackage{diagbox}

\newcommand{\datasetname}{PopCV}

\newcommand{\modelname}{LTAG}
\usepackage{comment}

\title{Translate the Beauty in Songs:\\Jointly Learning to Align Melody and Translate Lyrics}

\author{Chengxi Li* \\
Zhejiang University \\
\texttt{chengxili@zju.edu.cn} \\
\And
Kai Fan*\\
Alibaba DAMO Academy \\
\texttt{k.fan@alibaba-inc.com} \\
\And
Jiajun Bu\\
Zhejiang University \\
\AND
Boxing Chen \\ 
\texttt{chenboxing@gmail.com}\\
\And
Zhongqiang Huang \\
Alibaba DAMO Academy \\
\And
Zhi Yu\\
Zhejiang University \\
}
\begin{document}
\maketitle
\begin{abstract}
Song translation requires both translation of lyrics and alignment of music notes so that the resulting verse can be sung to the accompanying melody, which is a challenging problem that has attracted some interests in different aspects of the translation process. In this paper, we propose \underline{L}yrics-Melody \underline{T}ranslation with \underline{A}daptive \underline{G}rouping (\modelname), a holistic solution to automatic song translation by jointly modeling lyrics translation and lyrics-melody alignment. It is a novel encoder-decoder framework that can simultaneously translate the source lyrics and determine the number of aligned notes at each decoding step through an adaptive note grouping module. To address data scarcity, we commissioned a small amount of training data annotated specifically for this task and used large amounts of augmented data through back-translation. Experiments conducted on an English-Chinese song translation data set show the effectiveness of our model in both automatic and human evaluations.\footnote{The audio and score samples can be found at \href{LTAG2023.github.io}{\texttt{LTAG2023.github.io}}}

% Lyrics translation, in fact, is a part of song translation which requires other important considerations such as alignment to the music notes so that the result is not just a piece of text of any length or number of syllables but a verse that can be sung to the accompanying melody.
% \textbf{need correction:Song translation itself is a way to climb up the Tower of Bable.}
% In this paper, we consider the problem of song translation in a holistic manner by jointly modeling the translation of the lyrics and the alignment to the melody. We propose a novel encoder-decoder framework that can simultaneously translate the lyrics and predict the appropriate alignment between the text tokens and music notes. 
% Our \underline{L}yrics-Melody \underline{T}ranslation with \underline{A}daptive \underline{G}rouping (\modelname) framework is an adaptive note grouping model that dynamically determines the number of aligned notes at each decoding step. 
% Because data in this domain are especially scarce, we use both larger amounts of back-translated data and smaller amounts of supervised data commissioned specially for this purpose. 
% Experiments on English-Chinese bilingual song data set show the effectiveness and the potential of our model both in automatic and human evaluations. Both the audio and score samples can be found at \href{LTAG.github.io}{\texttt{LTAG.github.io}}.

\end{abstract}
\input{intro}
\input{relate}

\input{method}
\input{experiments}

\section{Conclusion}
In this work, we propose \modelname, a lyrics translation model with lyrics-melody alignments that allows simultaneous generation of target text and alignment to the music notes. 
We propose an adaptive grouping method that fits in the auto-regressive translation process. 
To better train and evaluate our model, we also annotate a new song translation data set \datasetname~containing English and Chinese songs with their cover version in both languages and with lyrics-melody alignments. 
For training, we also employ back-translation to leverage the more abundantly available monolingual lyrics data with lyrics-melody alignments in a curriculum learning way. 
Evaluations with both the automatic and human metrics show that \modelname~is capable of producing natural, singable and enjoyable translation results.
% than strong baselines.

\clearpage
\section{Concerns of the Ethical Impacts}
This work develops a possible automatic method for song translation. 
Therefore, if we release our repository and data set, there is the potential of abusing to synthesize score sheets and texts, especially may cause copyright issues. 
Thus, we choose the dataset license: CC by-nc-sa 4.0.
In this paper, we thoroughly discuss strengths and shortcomings of our proposed model and perform a series of experiments to support them. Codes, model checkpoints and data set will be released upon acceptance after desensitization and compliance examination.
% \section{-------Remember to add Limitations in the comment to Senior Area Chair of the submission to ACL 2023-------}
\clearpage
\bibliography{anthology}
\bibliographystyle{acl_natbib}
\clearpage
\input{appendix}
\end{document}

%% file: intro.tex
\section{Introduction}
\label{sec:intro}
\begin{figure}
    \centering
    \includegraphics[width=0.5\textwidth]{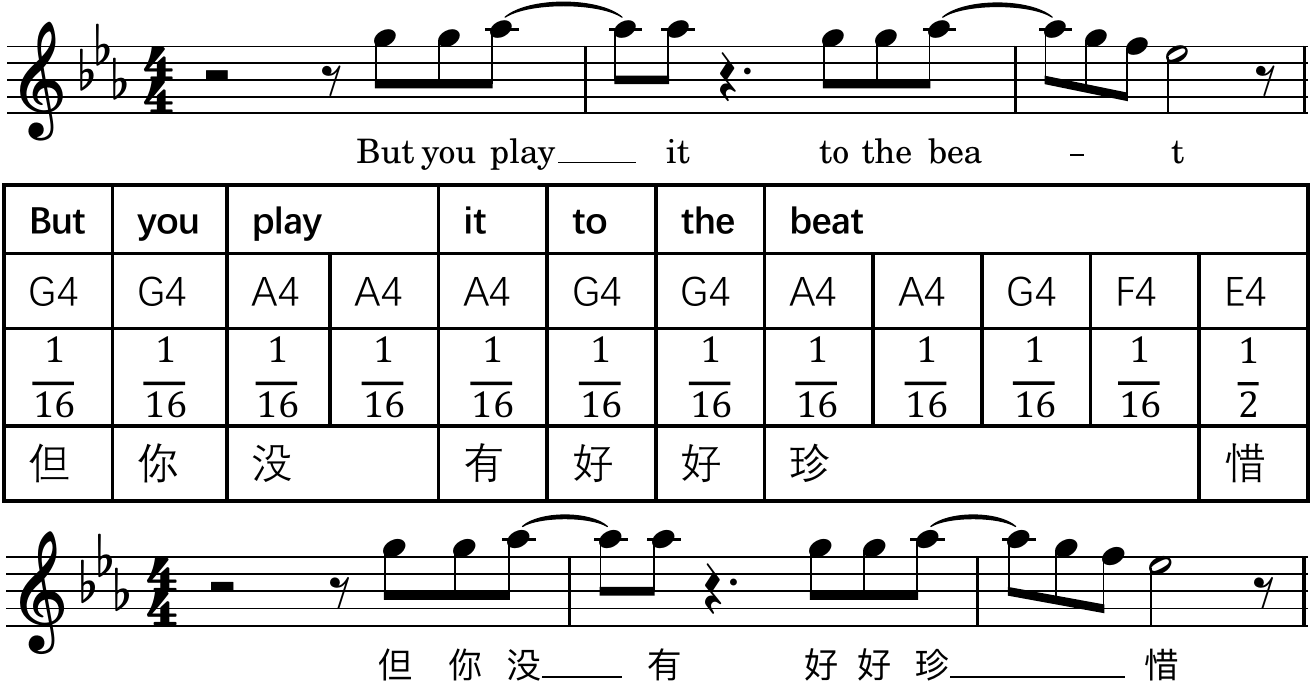}
    \caption{An example of the comprehensive translation for ``But you play it to the beat'' in \textit{Rolling In the Deep}. }
    % \modelname~aims to translate the lyrics and predict the number of aligned notes for each target token.
    \label{fig:task_exp}
\end{figure}

Song translation is a meaningful human endeavor to climb high Tower of Babel for inter-culture exchange.
% Song translation, a meaningful human endeavor for inter-culture exchange, 
Yet it has not received much attention in the natural language processing~(NLP) community despite the advancement of machine translation technologies, especially Neural Machine Translation (NMT)~\citep{nmt, vaswani2017attention, hassan2018achieving}, and the expanding interests of solving real-world problems using artificial intelligence techniques. Challenges include the lack of efficient means to collect parallel lyrics and alignment data, the difficulty of modeling the complex interaction between texts and melody and imperceptive evaluation of scores.
While closely related to text translation, song translation is a more involved task. In addition to the general considerations of word choice and word order in translation, human translators of songs need to have a mastery of cultural traditions and the poetic usage of both source and target languages. Furthermore, the translated lyrics need to be properly aligned with the melody, as shown in Figure \ref{fig:task_exp}, to maintain the intact beauty of the song, a factor that is indispensable in song translation~\citep{three_d_of_singability}.

% Song translation, a meaningful endeavor especially for the exchange of cultures, has not received enough attention in the natural language processing community partly because it seems to require interdisciplinary expertise in at least two areas, that of NLP and that of music, and partly because its sister problem, that of lyrics translation, seems to be readily subsumed in the general field of machine translation. 
% It is the task of translating the lyrics of a song from one language to another while maintaining the integrity between words and music. 
% While Neural Machine Translation (NMT)~\citep{nmt, vaswani2017attention, hassan2018achieving} has seen much success in domains such as news, translation technology for other forms of text, particularly those of art, is lacking, possibly with good reasons. 
% Song translation by humans is a more involved task than other text translations. 
% In addition to general translation considerations such as diction and word order, human translators of songs need to have a mastery of poetic usage of language and cultural traditions of both the source and the target. 
% Furthermore, the translated lyrics need to be re-aligned with the melody so that they can be sung appropriately. 

Researchers have explored Singing Voice Synthesis (SVS)~\citep{diffsinger, pndm, nsvb} to automate the vocal singing of songs
% , and proposed approaches to produce
% natural and accurate singing voice with realistic vocal timbre 
given the input lyrics and scores, which laid the foundation of convenient and perceptive evaluation and a prospective empirical usage of automatically generated songs. However, there is very few previous studies in the direction of Automatic Song Translation (AST). The sole work~\citep{gagast} we are aware of focuses on matching tones and rhythms for the translated target words for tonal languages, by imposing constraints during NMT inference. Their direct use of text translation models and strict mapping between notes and tokens, however, is unable to capture the more involved nature of song translation. While the number of notes provides an easy upper bound on the length of translation, the delicate alignment between lyrics and melody, as observed in \citet{interplay_lyrics_melody}, should not be dictated solely by simple rigid rules. 

% Various stages in this process have been the goals of automation. 
% For example, the vocal singing of songs has been extensively explored as a Singing Voice Synthesis (SVS) problem~\citep{diffsinger, pndm, nsvb}. 
% It is capable of producing natural and accurate singing voice with realistic vocal timbre given the input lyrics and scores. 
% Recent Automatic Song Translation (AST)~\citep{gagast} research focuses on the translation quality of the lyrics by imposing special constraints on the decoder, for example, on the length or on the tones of words if the target language is tonal. 
% Although these special considerations ensure the length of the lyrics does not exceed that of the music, the entire process is still more text translation than song translation. 
% For song translation, \citet{three_d_of_singability} has shown that singability is an indispensable factor a crucial component of which is the alignment between the lyrics and the notes. 
% In this area, both rigid and flexible alignments have been proposed. 
% \citet{gagast} ) is a one-to-one mapping between notes and words, while \citet{interplay_lyrics_melody} is a flexible alignment which works more reasonably in a real situation. 
% Although the number of notes for one verse of lyrics is an easy upper bound of the length of the translation, more can be utilized for better results. 
% In the auto-regression translation model, the correct number of aligned notes can be dynamically determined and assigned to the corresponding target words in the lyrics.

In this paper, we propose Lyrics-Melody Translation with Adaptive Grouping (\modelname), the first comprehensive solution to the AST problem, by jointly modeling of both lyrics translation and lyrics-melody alignment within the transformer-based encoder-decoder framework. \modelname~incorporates both lyrics and melody in an end-to-end manner and employs an adaptive grouping module to explicitly model the alignment between lyrics and melody. To facilitate training, we produce the first (Chinese-English) bilingual lyrics-melody alignment data set. To address the data scarcity problem, we also generate a large amount of bilingual lyrics-melody data through back-translation of monolingual lyrics-melody alignment data, which is used together with the high quality manual annotations through a curriculum training strategy. Our experiments show that songs translated by \modelname~are both faithful to the original lyrics and singable to the melody, as measured by both automatic metrics and human judges majoring in music. Main contributions of this work are as follows:

% To tackle these challenges, on top of the transformer encoder-decoder framework, we design a novel adaptive grouping module for implicit alignment inference. 
% Our model of Lyrics-Melody Translation with Adaptive Grouping (\modelname) is the first co-translation framework to approach the AST problem in a more comprehensive manner as shown in Figure~\ref{fig:task_exp}. 
% For this purpose, we produce the first bilingual lyrics-melody alignment data set. 
% In addition to this supervised set, we also use the popular technique of back translation to add more monolingual lyrics-melody alignment data.
% A curriculum learning way is used to balance the usage between back-translated data and human-annotated data. 
% Songs translated by \modelname~are both faithful to the original lyrics and adapted to the melody. 
% In addition to automatic metrics, we also conduct human evaluations on the results by annotators majoring in music. The main contributions of this work are:

%\begin{enumerate}

\noindent
\textbf{(1)}
%\item 
We propose the first joint lyrics translation and lyrics-melody alignment framework \modelname~to solve the AST task in a comprehensive manner. 

\noindent
\textbf{(2)}
%\item 
We design an adaptive grouping method for monotonic lyrics-melody alignment prediction that helps achieve high-quality lyrics translation and to provide flexible and reasonable lyrics-to-melody alignments in the same auto-regressive process.

\noindent
\textbf{(3)}
%\item 
We produce the first bilingual lyrics-melody alignment data set that will be released publicly to facilitate further research in this field. 
We also leverage the back-translation and the curriculum learning strategy to boost performance.

\noindent
\textbf{(4)}
%\item 
Our experiments show that \modelname~outperforms baselines by a notable margin. 
Human evaluations indicate that our proposed flexible alignments together with lyrics translation achieves satisfying song translation results.

%\end{enumerate}

%% file: relate.tex
\section{Related Works}

\begin{figure*}[t]
    \centering
    \includegraphics[width=0.99\textwidth]{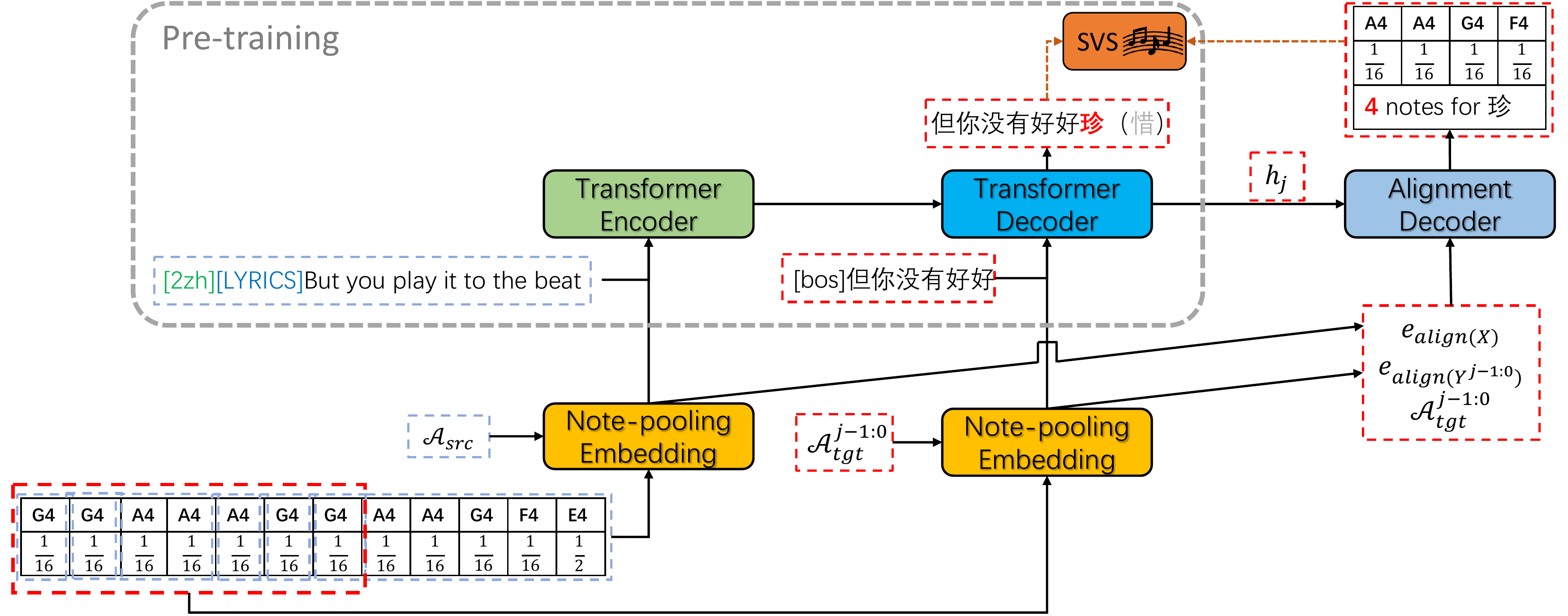}
    \caption{The overall overview of our proposed architecture, illustrated at the $j$-th decoding step. The transformer decoder will output the target token and the alignment decoder will derive the number of aligned notes.}
    \label{fig:model}
\end{figure*}

%\subsection*{Verse and Song Translation}
\noindent
\textbf{Lyrics and Song translations} have recently drawn attention from the NLP community. 
Automatic lyrics translation has been approached by rule-based methods, statistical machine translation methods, finite-state methods with rhythmic and lexical constraints~\citep{spanish_verse, Manurung2004AnEA, He_Zhou_Jiang_2012}, and more recently by neural methods~\citep{ghazvininejad-etal-2016-generating,ghazvininejad-etal-2017-hafez, ghazvininejad-etal-2018-neural}. 
Traditional song translation research has made progress in terms of lyrics translation and lyrics-melody alignment through linguistic knowledge~\citep{interplay_lyrics_melody,low_2003,low2008translating,low_2022,three_d_of_singability,trans_of_music}. 
Often, the object of these research was artificial songs. 
These methods pursue lyrics-melody alignment and lyrics translation in separate tracks. 
% \citet{gagast} pursue song translation as a type of Constrained Text Generation. 
\citet{gagast} pursue song translation as a type of Constrained Text Translation. 
Previous works~\citep{hokamp-liu-2017-lexically,lakew-etal-2019-controlling,li-etal-2020-rigid,zou_controllable} on constraining the decoding process are shown to be effective in performance and convenient in implementation. 
Others that impose constraints during training, such as adding format embedding ~\citep{li-etal-2020-rigid}, introducing special tags and rescoring length control ~\citep{lakew-etal-2019-controlling,saboo-baumann-2019-integration}, are data-driven methods and show good performance. 
In this paper, we propose the lyrics translation model with lyrics-melody alignments for domain shift and length control, and overcome the problem of domain mismatch and data sparsity by using monolingual data. 

%\subsection*{Lyrics Generation and Alignment Prediction}
\noindent
\textbf{Lyrics Generation with Alignment Prediction}, one of the most important tasks in automatic song production, has received much attention recently. 
Most of the current works~\citep{lee-etal-2019-icomposer,Chen2020MelodyConditionedLG,songmass,telemelody,ai_lyricist,xue-etal-2021-deeprapper} adopt the sequence generation method, but with different objectives. 
Some constrain rhythmic alignment, others theme and target genre. 
Other works~\citep{songmass,telemelody} apply the attention mechanism and find the lyrics-melody alignment via dynamic programming on the attention weights matrix. 
This method sometimes results in non-monotonic output. 
% and requires relatively large amounts of data in training
% Most importantly, perhaps, is that its alignment component is akin to a rule-based fixed constraint rather than an integrated unit that learns the dynamic alignment during training. 
Most importantly, it seems that their alignment component is akin to a post-processing module rather than an integrated unit that learns the dynamic alignment such that the lyrics generation is constrained. % and influences the decoding during inference.
In our proposal, we take advantage of the monotonic nature of the lyrics-melody alignments and design a light neural network for alignment prediction in parallel to the translation process.

%% file: method.tex
\section{Methodology}
In this section, we first describe the \modelname~as shown in Figure \ref{fig:model}. 
Then, we detail the adaptive grouping method for alignment prediction and explain how we adapt back-translation for the AST task. 

\subsection{Overall Architecture}

\begin{figure*}[t]
    \centering
\subfigure[The note-pooling embedding layer]{
    \label{fig:align_enc}
    \includegraphics[width=0.34\textwidth,clip=true]{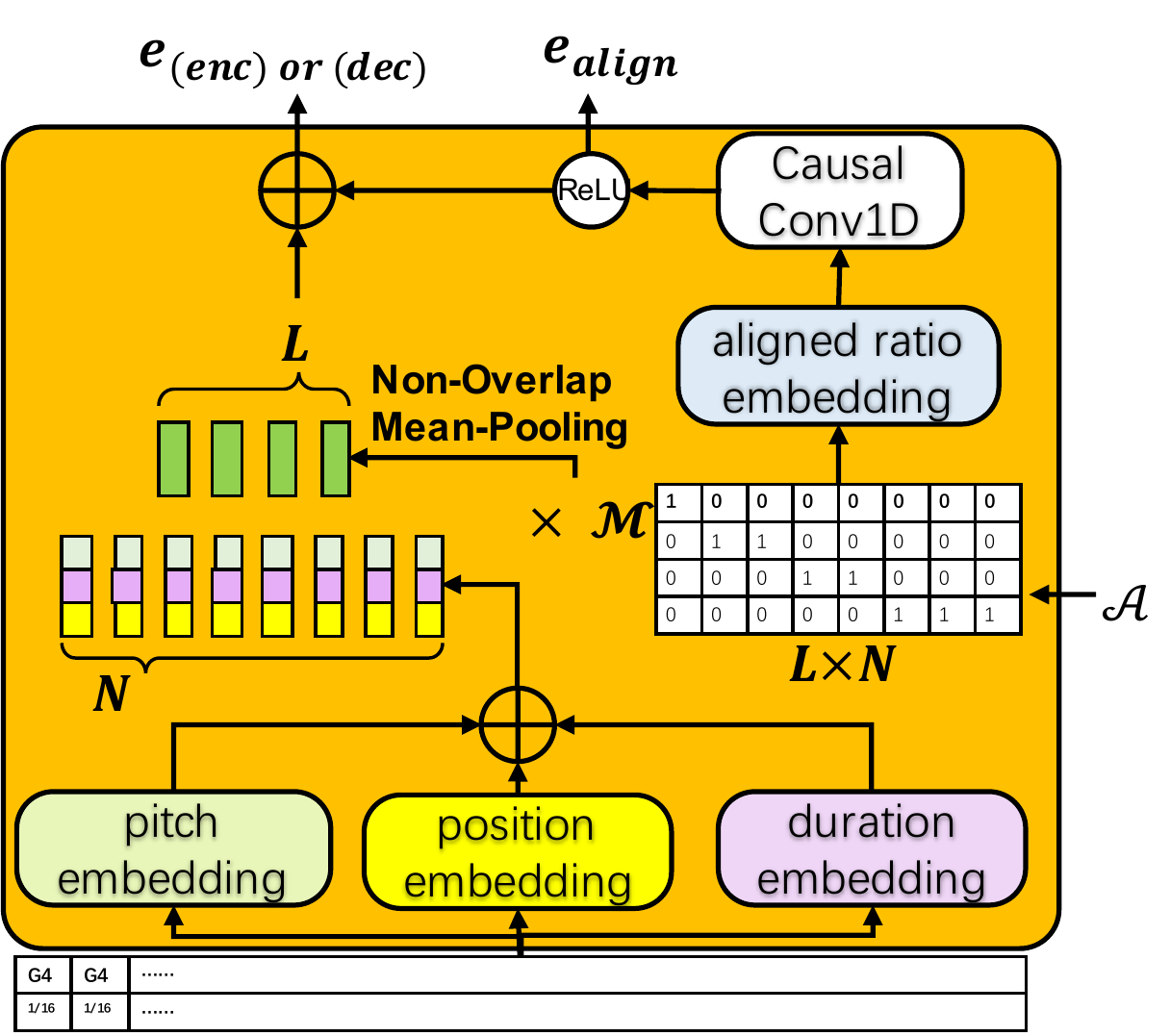}
}
\subfigure[The alignment decoder]{
    \label{fig:align_dec}
    \includegraphics[width=0.31\textwidth,clip=true]{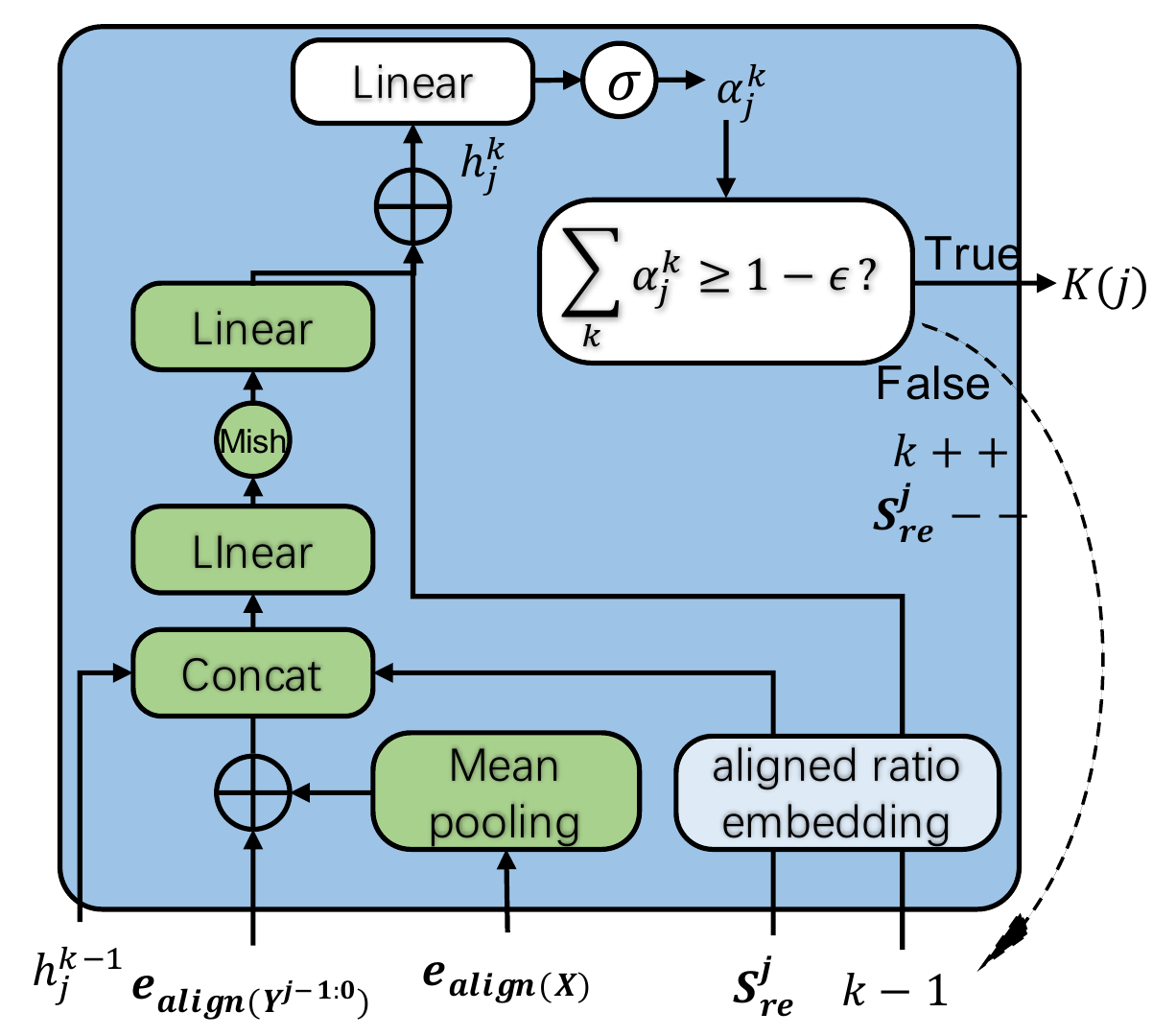}
}
\subfigure[Adaptive grouping process]{
    \label{fig:act_gp}
    \includegraphics[width=0.28\textwidth,clip=true]{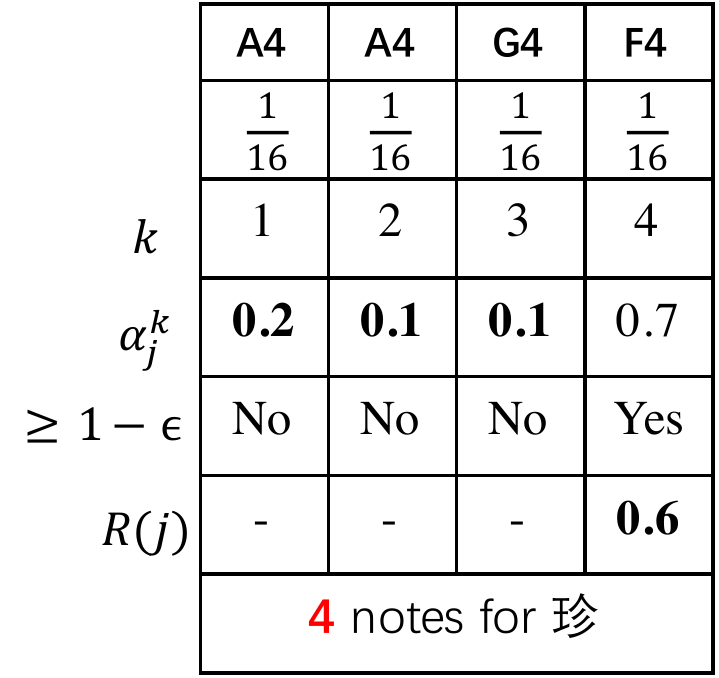}
}
\caption{(a) The note-pooling embedding encodes both the note sequence and the alignment information. (b)(c) The alignment decoder computes the number of aligned notes from halting distribution.}
\label{fig:enc_dec_act}
\end{figure*}

% We design an auto-regressive translation architecture that performs jointly lyrics translation and lyrics-melody alignment. 
We design an auto-regressive translation architecture that jointly performs lyrics translation and lyrics-melody alignment prediction.
As shown in Figure \ref{fig:model}, it consists of a transformer-based encoder-decoder pack for lyrics translation, two note-pooling embedding layers that embed and do pooling for notes and alignments, and an alignment decoder. 
The transformer encoder-decoder is pre-trained with a denoising auto-encoder ~\citep{bart} and for the translation task as in \citet{gagast}. 
During pre-training, two prefix tokens indicating the translation direction and the text domain are prepended to the source input. 
The note-pooling embedding layers shown in Figure~\ref{fig:align_enc} is a module that processes the melody information. 
The alignment decoder shown in Figure~\ref{fig:align_dec} is based on our adaptive notes grouping method that dynamically predicts the number of notes to align to a token during auto-regressive decoding.

\subsection{Note-Pooling Embedding}
\label{sec:note_pooling}

The note-pooling embedding layer takes the notes and alignments as input, and outputs the pooled note embedding and alignment embedding. 
The input note sequence consists of MIDI pitch and duration of each note. 
%Each note has one of the 4 different duration: quarter note par bar, half note per bar, eighth-note per bar, or two quarter-notes per bar. 
The MIDI pitch and duration can be represented as embedding $e_{midi}$ and $e_{dur}$ respectively. 
We define the $i$-th note embedding:
\begin{equation}
\label{eq:note}
    \mathbf{e}_{note}^i = \mathbf{e}_{midi}^i + \mathbf{e}_{dur}^i + \mathbf{e}_p^i
\end{equation}
where $\mathbf{e}_p^i$ is the positional embedding. 

We apply non-overlapping mean-pooling on the note embedding sequence according to the alignment information. 
Specifically, the embeddings of the consecutive notes that align to the same token are averaged. 
Mathematically, the alignment information $\mathcal{A}$ is represented as a binary matrix $\mathbf{M} \in \{0,1\}^{L \times N}$, where $L$ and $N$ denote the sequence length of the tokens and notes. 
$\mathbf{M}_{ji}=1$ if the $i$-th note is aligned to the $j$-th token. 
We use $\mathbf{M}$ to efficiently calculate the non-overlapping mean-pooling via matrix multiplication, denoted the result as melody embedding $\mathbf{e}_{md}$. 
\begin{equation}
\label{eq:md_embed}
    \mathbf{e}_{md} = \text{Non-Overlap-Mean-Pool}(\mathbf{e}_{note}, \mathbf{M})
\end{equation}
The kernel size of this operation is not fixed but varies according to the row sum of $\mathbf{M}$. 
The detailed calculation can refer to Appendix \ref{appendix:pool_mat}. 

Because lyrics-melody alignments are monotonic, we encode the alignment more succinctly by calculating the cumulative sum of the number for aligned notes:
\begin{equation}
\label{eq:cumsum}
    \mathbf{s} = \text{CumSum}(\text{RowSum}(\mathbf{M})) 
\end{equation}
where $\mathbf{s}$ is a vector of length $L$. 
$s^j / N$ then represents the alignment ratio for each aligned note. 
We next quantize the cumulative alignment ratios by grouping them into equal-size bins over the range $(0, 1]$, and introduce a set of embedding vectors $\mathbf{E}_{ratio}$ to represent each bin. 
Finally, the alignment embedding is calculated as follows.
\begin{align}
\label{eq:align}
    \mathbf{e}_{align}^j = f(\mathbf{E}_{ratio}(s^j / N))
\end{align}
where $f(\cdot)$ is a simple non-linear layer of causal 1D convolution with ReLU activation, and the number of bins is a hyper-parameter. 
The motivation is to implicitly constrain the translation by the number of aligned notes.

The melody embedding and alignment embedding are summed and then added to the original transformer encoder or decoder input. 
\begin{equation}
\label{eq:embed}
    \mathbf{e}_{\text{enc(dec)}} = \mathbf{e}_{token} + \mathbf{e}_p + (\mathbf{e}_{md} + \mathbf{e}_{align})
\end{equation}
As calculated in Eq.~(\ref{eq:md_embed}), each melody embedding corresponds to one token. 
In addition, the causal convolution implies that the alignment embedding tensors also have the same length as the text tokens and guarantees each alignment embedding only observes previous ratio embeddings in an auto-regressive manner. 
It means that on the decoder, this layer can fit perfectly in the teacher-forcing training. 

%The alignment embeddings from source then go through a pooling layer to form a global reference representation and will be added to the target alignment embeddings in the alignment decoder.

\subsection{Alignment Decoder}
\label{sec:alignment_decoder}

Inspired by the Adaptive Computation Time (ACT) \citep{act}, we propose the \textbf{adaptive grouping} module to model lyrics-melody alignment. 
As shown in Figure \ref{fig:align_dec} and \ref{fig:act_gp}, this module predicts how many consecutive notes should be assigned to the current token. 

For $1 \leq j \leq L_Y$, let $y_j$ be the $j$-th target token and $\mathbf{h}_j$ be the corresponding hidden state of the last transformer decoder layer. 
Suppose previous tokens $y_{j-1:0}$ have been aligned to the first $n-1$ notes, we define the following adaptive grouping process by iterating over index $k$ (starting from 1) to derive the number of notes aligned to $y_j$.
\begin{align*}
    \mathcal{S}_{re}^j &= N -s^{j-1}_{tgt} \\
     \mathbf{h}_j^0 &= \mathbf{h}_j  \\
     \mathbf{h}_j^k &= g(\mathbf{h}_j^{k-1}, \mathbf{e}_{align(X)}, \mathbf{h}_{align(y_{j-1:0})}, \mathcal{S}_{re}^j, k-1) \\
     \alpha_j^k &= \sigma(\text{Linear}(\mathbf{h}_j^k))
\end{align*}
where $\mathbf{e}_{align(X)}$ and $\mathbf{e}_{align(y_{j-1:0})}$ are the alignment embeddings of the full source input and the partial target input respectively, and $s^{j-1}_{tgt}$ is $j$-th element of vector $\mathbf{s}$ in Eq.~(\ref{eq:cumsum}). 

We first calculate the residual number of unaligned notes at the current decoding step $j$ as $\mathcal{S}_{re}^j$. 
$\mathbf{e}_{align(X)}$ is fed into an average pooling layer to obtain a single vector, making it always possible to be additive with $\mathbf{e}_{align(y_{j-1:0})}$ of variable length. 
For all the inputs, we apply a multi-layer network $g(\cdot)$ shown in green in Figure \ref{fig:align_dec}. 
Eventually, the sigmoid function $\sigma(\cdot)$ outputs the halting probability $\alpha_j^k$ of the intermediate step. 
The summation of these probabilities represent the likelihood that the current $k$ notes are aligned to the target token $y_j$. 

Given a hyper-parameter $\epsilon$ as a small float number (\emph{e.g.}, 0.01), if $\sum_k \alpha_j^k < 1-\epsilon$, the adaptive grouping process will continue and re-calculate by incrementing $k$ and decrementing $\mathcal{S}_{re}^j$.
Otherwise, the aligning process halts, and the alignment decoder outputs the number of aligned notes $K(j)$.
\begin{equation}
\label{eq:Kj}
    K(j) = \underset{K}{\mathrm{argmin}} \left\{\sum_{k=1}^K \alpha_j^k \geq 1-\epsilon \right\}
\end{equation}
A positive $\epsilon>0$ guarantees that $K(j)\geq1$, \emph{i.e.}, at least one note is aligned. 
To define the halting probabilities of $K(j)$ aligned notes, we introduce the remainder $R(j) = 1-\sum_{k=1}^{K(j)-1} \alpha_j^k$. 
%
%\begin{equation}
%    R(j) = 1-\sum_{k=1}^{K(j)-1} \alpha_j^k
%\end{equation} 
%
In this way, $\alpha_j^k$ and $R(j)$ can be valid probability distributions. 
Figure \ref{fig:act_gp} is an example of how the adaptive grouping works.

In the labeled alignment data, the ground truth of the number of aligned notes for each target token is available, denoted as $\Delta_j$. 
Instead of minimizing the ponder cost $\sum_j K(j) + R(j)$ as in ACT \citep{act}, we optimize the following adaptive grouping loss $L_G$, which could naturally upper bound the token-wise ponder cost via $\Delta_j$. 
\begin{equation*}
\begin{array}{rl}
    L_G = & \left| \sum_j K(j) - N \right| + \sum_j \left|K(j) - \Delta_j\right| \\
    \approx &\left|\sum_j \left(K(j) - (1 - R(j))\right) - N \right| \\
    & + \sum_j \left|K(j) - (1 - R(j)) - \Delta_j\right|
\end{array}
\end{equation*}
The variable $K(j)$ is discontinuous with respect to the halting probabilities, so we use $1-R(j)$ in the approximation to make the loss differentiable (more analysis in Appendix \ref{appendix:group_loss}). 
Additionally, because tokens aligned to more than one notes are infrequent, we add upweighting to the alignment loss of such tokens for model calibration.
\begin{equation*}
\begin{array}{rl}
    L_G = & \left|\sum_j (K(j) - 1 + R(j)) - N\right| \\
    & + \sum_j (\left|K(j) - 1 + R(j) - \Delta_j\right|\cdot w_j)
\end{array}
\end{equation*}
where $w_j=1$ if $\Delta_j = 1$ and $w_j>1$ is a hyper-parameter if $\Delta_j > 1$.

\subsection{Back Translation with Alignments}
\label{sec:bta}
Although a data set of a few thousand verses with human translation and annotated with alignment information is useful, its quantity is limited.
We therefore adopt the widely used back-translation method \cite{backtrans} to generate more training data. 
We crawl the web for more available monolingual song data with alignments and build another pre-trained lyrics translation model with length control that is used to back translate the monolingual data into the source language. 
The length control ensures that the number of tokens is the same as the number of notes after which a one-to-one source-side alignment can be generated. 
This way, we obtain a comparatively larger data set with noise on the source side but still accurate information on the target side. 

Because the back-translated data are much larger than the human annotated one, we in practice design our data loader by following a curriculum learning way. 
Initially, the augmented data from back-translation will be mixed with up-sampled the real data from human annotation. 
In each training epoch, we gradually down-sample the augmented data to raise the ratio of annotated data in the batch. 
A visualization of the data sampling scheduler is in Figure~\ref{fig:bt_curriculum} (See Appendix \ref{appendix:bt_cl}).

\subsection{Training and Inference}

After the pre-training stage, we will optimize the whole model by jointly minimizing the loss from the task of lyrics translation and the task of alignment prediction. 
% Note that the SVS model is pre-trained and will not participate the training of song translation. 
Note that the SVS model is pre-trained and only used for evaluation. 
The overall loss is thus:
\begin{equation*}
    L_{joint} = \sum_{j=0}^{L_Y} \log P(y_j|y_{j-1:0},X) + \beta \cdot L_G
\end{equation*}
where $\beta$ is a hyper-parameter to balance the importance between the two tasks.
The inference follows the standard beam search for auto-regressive decoding, while only the last generated token and its corresponding notes should be specially taken care of. 
Details can be found in Appendix \ref{appendix:infer}.

%% file: experiments.tex
\section{Experiments}
In this section, we describe the experiment setup, results and analysis on Chinese$\leftrightarrow$English song translation.

\begin{table*}[t]
    \centering
    \begin{tabular}{l|c|c|c|c|c|c}
    \hline
    \multirow{2}{*}{Models} & \multicolumn{2}{c|}{MOS-T} & \multicolumn{2}{c|}{MOS-S} & \multicolumn{2}{c}{MOS-Q} \\
    \cline{2-7}
    & En$\rightarrow$Zh & Zh$\rightarrow$En & En$\rightarrow$Zh & Zh$\rightarrow$En$^\dagger$ & En$\rightarrow$Zh & Zh$\rightarrow$En$^\dagger$ \\
    \hline\hline
    Human Ref. & 3.83 $\pm$ 0.06 & 4.11 $\pm$ 0.05 & 3.92 $\pm$ 0.07& \multirow{8}{*}{\diagbox[height=40pt, width=0.05\textwidth]{}{}} & 3.90 $\pm$ 0.06 &\multirow{8}{*}{\diagbox[height=40pt, width=0.05\textwidth]{}{}}\\
    \cline{1-4} \cline{6-6}
    GagaST & 3.66 $\pm$ 0.06 & 3.72 $\pm$ 0.05 & 3.49 $\pm$ 0.10 & & 3.65 $\pm$ 0.05 & \\
    \cline{1-4} \cline{6-6}
    \modelname-cls  & 3.66 $\pm$ 0.05& 3.79 $\pm$ 0.05 & 3.58 $\pm$ 0.07& & 3.62 $\pm$ 0.05& \\
    ~~~ only bt & 3.69 $\pm$ 0.05 & 3.80 $\pm$ 0.04 & 3.53 $\pm$ 0.09 & & 3.63 $\pm$ 0.05&\\
    ~~~ w/o bt & 3.64 $\pm$ 0.05 & 3.30 $\pm$ 0.05 & 2.16 $\pm$ 0.05 & & 3.14 $\pm$ 0.04 &\\
    \cline{1-4} \cline{6-6}
    \modelname  & 3.71 $\pm$ 0.05& 3.85 $\pm$ 0.05 & 3.68 $\pm$ 0.05&  & 3.69 $\pm$ 0.04&\\
    ~~~ only bt & 3.71 $\pm$ 0.05 & 3.80 $\pm$ 0.05 & 3.58 $\pm$ 0.07 & & 3.65 $\pm$ 0.04&\\
    ~~~ w/o bt  & 3.69 $\pm$ 0.05 & 3.28 $\pm$ 0.04 & 3.63 $\pm$ 0.07 & & 3.67 $\pm$ 0.04&\\
    % \midrule
    % \modelname~w/o bt  & & & & & & \\
    % \modelname~only bt & & & & & & \\
    % \modelname~+ bt  & & & & & & \\
    \hline
    \end{tabular}
    \caption{The Mean Opinion Score in translation intelligibility and naturalness~(MOS-T), singability~(MOS-S) and overall quality~(MOS-Q) with 95\% confidence intervals. The translation direction with $^\dagger$ means that audio samples of the translated song for evaluation are generated with the voice synthesis model that is not trained for that target language. So those results are presented in Appendix \ref{appendix:zh-en} and for reference only.}
    %带*的结果仅供参考
    \label{tab:subjective}
\end{table*}

\subsection{Experimental Settings}

\subsubsection*{Data Sets}

Since there is no publicly available data set with high quality parallel lyrics translation and lyrics-melody alignments, we collect and annotate a data set PopCV (Pop songs with Cover Version) containing both Chinese songs with their English cover version and English ones with Chinese cover version. 
Since there are no industry standards or published precedence in annotating such a data set, we design an annotation procedure which is time-saving and easy for annotators to carry out. 
First, we collect the score sheet files of songs from score websites\footnote{\scriptsize{\url{https://www.musescore.com} and \url{https://wwww.midishow.com}}}. 
Then the annotators add lyrics to notes according to how songs are sung in the original and its cover version as conventions\footnote{\scriptsize{\url{https://lilypond.org} and \url{https://musescore.org/howto}}} suggest. 
We then export the annotated music score files in \texttt{.musicxml} format and automatically extract lyrics and their aligned notes. 
Please refer to Appendix \ref{appendix:data} for details. 

% \begin{figure*}[htbp]
%     \centering
%     \includegraphics[width=0.99\textwidth]{figures/data_annotation.pdf}
%     \caption{An illustration of data collection and annotation process.}
%     \label{fig:data_anno}
% \end{figure*}
For the data used in back-translation, we use LMD\footnote{\scriptsize{\url{https://github.com/yy1lab/Lyrics-Conditioned-Neural-Melody-Generation}}}~\citep{LMD} for English songs with alignments to melody and a data set crawled from Changba App for Chinese songs. 
We first pre-train two lyrics translation models with length control, one in each direction, and then translate the above two data sets.
The translated lyrics are one-on-one aligned to the notes. 
Two sets of back-translated data are used for training only while testing is done on real data with human annotations. 
An overview of the data is in Table \ref{tab:dataset_stat}. 
We will release the code and the human annotated data set upon acceptance. 

\begin{table}[t]
    \centering
    \setlength{\tabcolsep}{2pt}
    \begin{tabular}{l|c|c|c|c}
    \hline
         & Lang & Songs & Lyrics & Source\\
    \hline
     LMD & En & \diagbox[]{}{} & 152,991 & Back Translation\\
    \hline
     Changba & Zh & \diagbox[]{}{} & 542,034 & Back Translation\\
    \hline
     PopCV & En,Zh & 79 & 2,959 & Annotation\\
    \hline
     testset & En,Zh & 25 & 629 & Annotation\\
    \hline
    \end{tabular}
    \caption{Statistics of datasets in our experiments}
    \label{tab:dataset_stat}
\end{table}

\subsubsection*{Evaluation Metrics}

The most convincing evaluation of how our model works is whether the translated songs can be sung, understood, and, most importantly, enjoyed. 
Thus, we follow \citet{songmass}~and show annotators the resulting score of the song with translated lyrics. 
To verify the singability in the end-to-end manner, we additionally use an open-source Chinese singing voice synthesis (SVS)  model~\citep{diffsinger} to supply the annotators with an actual audio rendition of the songs for more intuitive feeling.

We randomly select 20 verses from the test set and show the music sheets and synthesized singing voice~(see Appendix \ref{appendix:data}) of each translated verse to five annotators.
For automatic evaluations, we use sacreBLEU\footnote{\scriptsize{\url{https://github.com/mjpost/sacrebleu}}}.  
For translation intelligibility, naturalness, singability and overall quality evaluation, we use mean opinion score~(MOS) in human evaluations, referred to MOS-T, MOS-S and MOS-Q. 
In evaluating the alignments, the traditional AER does not apply here because in addition to machine-produced alignments, the target translation is also machine-produced. 
Instead, we propose an Alignment Score (AS) that calculates the weighted intersection over ground truth (IOG) of the empirical probability density between the predicted and the true alignments:
%改成重合对应音符数的期望值和真实音符数的期望值之间的比值
%
\begin{equation}
    % \text{AS} = \frac{\sum_{k} \min(\text{freq}_{pred}^k, \text{freq}_{gt}^k) * k)}{\sum_{k} \text{freq}_{gt}^k * k }
    \text{AS} = \frac{\sum_{k}\min(\text{freq}_{pred}^k/F_{pred}, \text{freq}_{gt}^k/F_{gt}) * k)}{\sum_{k} (\text{freq}_{gt}^k/F_{gt}) * k) } 
\end{equation}
where $k$ represents the number of aligned notes, and $F = \sum_{k} \text{freq}^k$. 

\begin{figure*}[t]
    \centering
% \subfigure[GT]{
%     \includegraphics[width=0.23\textwidth,clip=true]{figures/align_hist.pdf}
% }
\subfigure[GagaST]{
    \includegraphics[width=0.31\textwidth,clip=true]{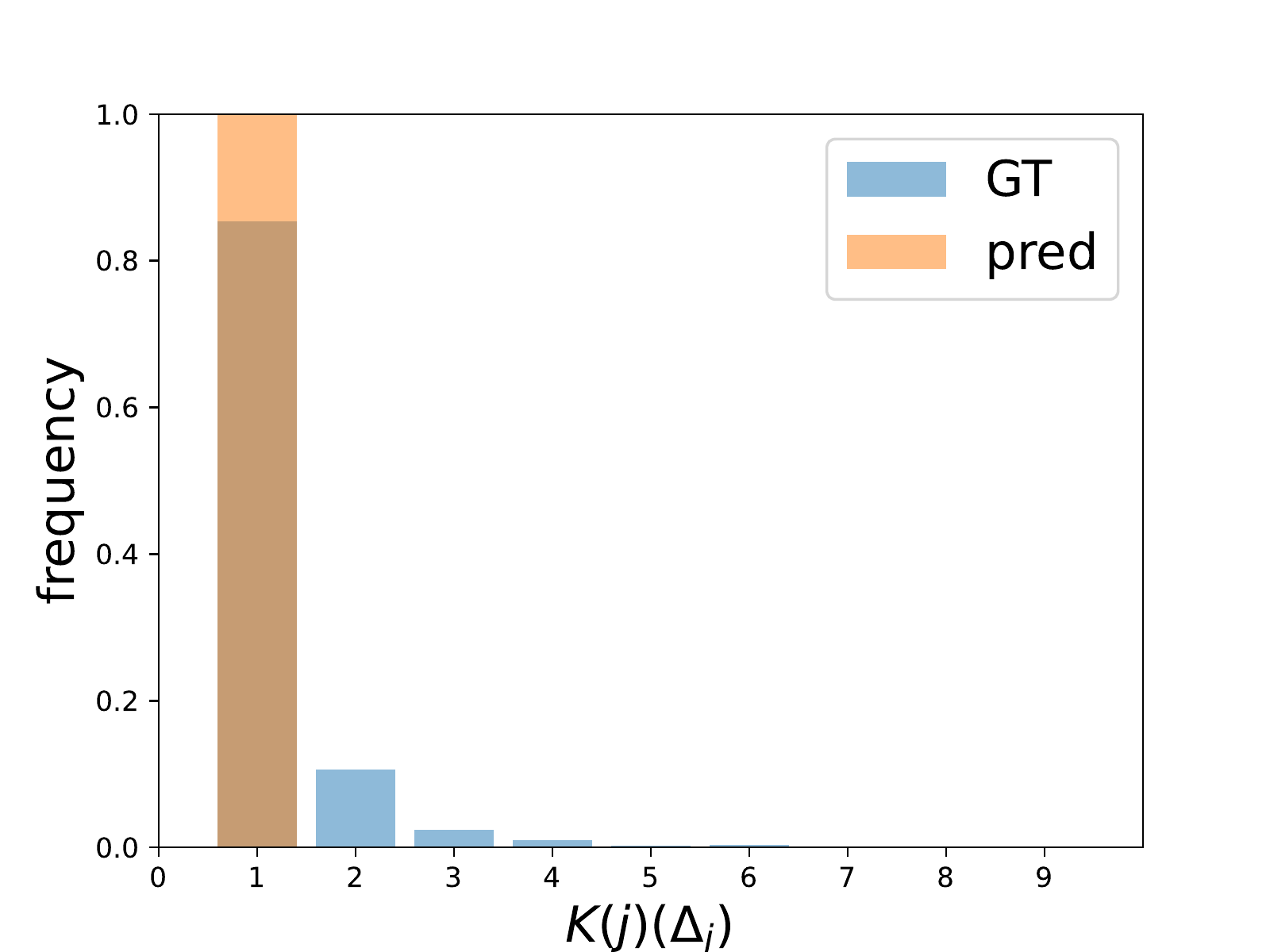}
}
\subfigure[\modelname-cls]{
    \includegraphics[width=0.31\textwidth,clip=true]{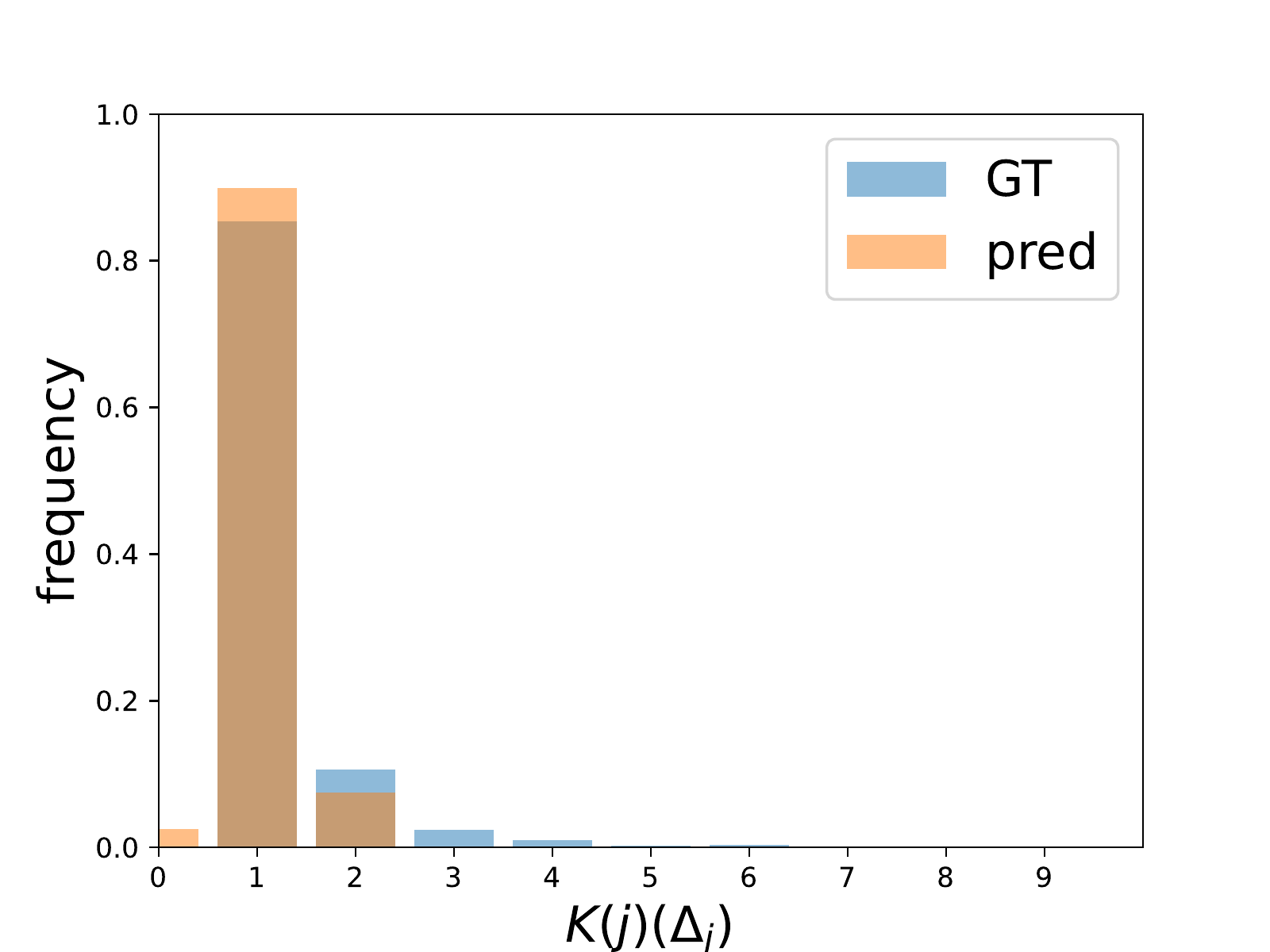}
}
\subfigure[\modelname]{
    \includegraphics[width=0.31\textwidth,clip=true]{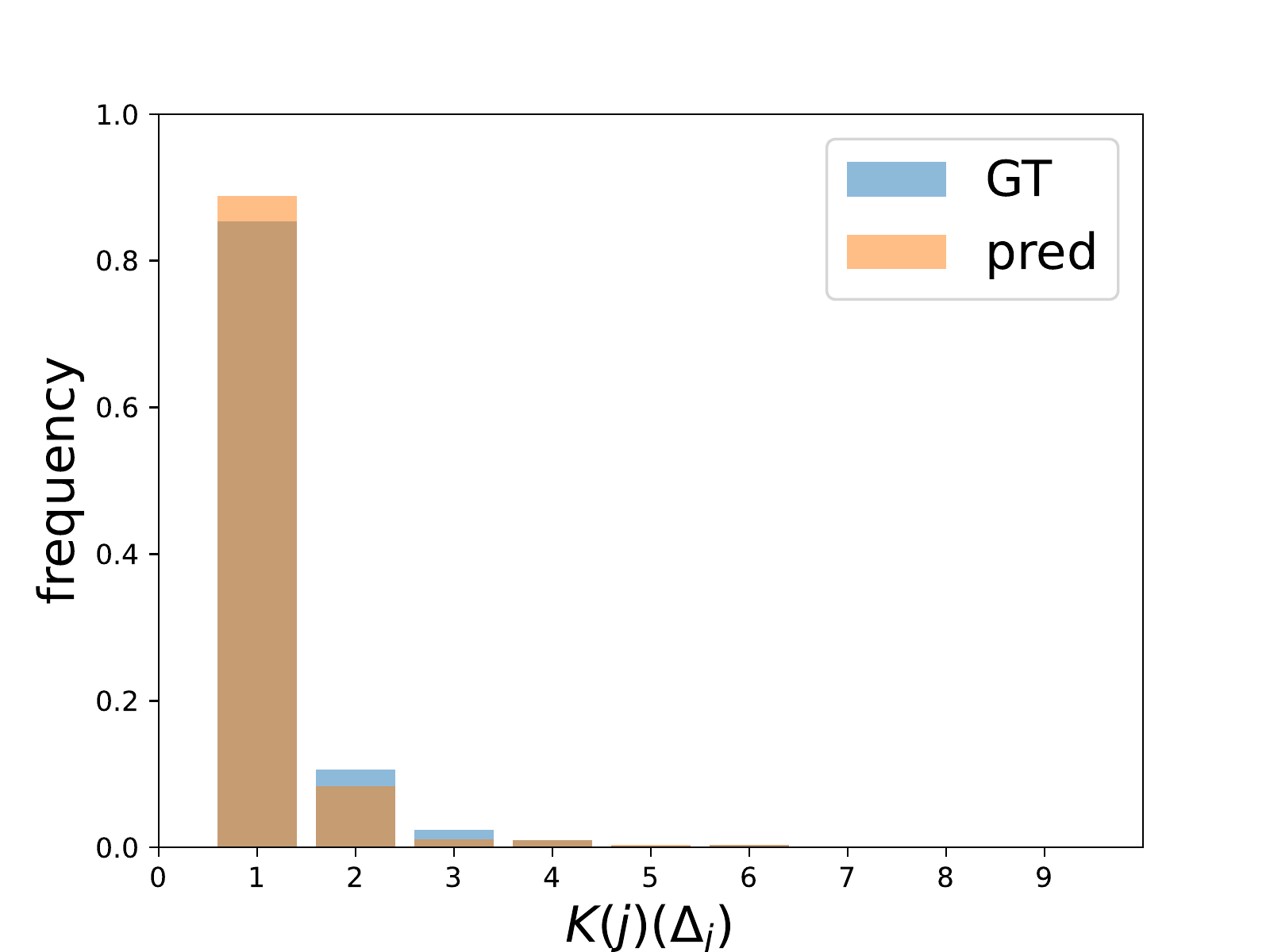}
}
\caption{The overlapped histograms of ground truth alignments and predicted alignments on En$\rightarrow$Zh test set.}
\label{fig:align_hist}
\end{figure*}

\subsubsection*{Model Configurations}

The token embeddings of the Transformer encoder and decoder have dimension of 256 and are shared.
In the note-pooling embedding layer, the size of the lookup table for MIDI pitch and duration type are set to 128 and 31. 
The halting hyper-parameter epsilon $\epsilon$ for the adaptive grouping process is 0.05. 
$w_j$ is 5 when $\Delta_j > 1$, and $\beta$ is 0.8 in the joint loss $L_{joint}$. 
The beam size during decoding is 5.

The \modelname~model is pre-trained on the WMT data and the crawled lyrics data, including the parallel and the monolingual corpora.
% , with 4 Tesla A100 GPUs and 20480 max tokens per batch. 
% For the co-translation training, we use one Tesla V100 GPU with 4096 max tokens per batch.
The sampling ratio scheduling of augmented data and annotated data are described in Appendix \ref{appendix:bt_cl}. 

For voice synthesis, we convert the Chinese lyrics into phonemes by \emph{pypinyin}~\citep{ren2020deepsinger} and set the hop size and frame size to 128 and 512 for the sample rate of 24kHz. 
Pitch inputs to the SVS model are all re-tuned to the range between $A3$ and $C5$ in C major. 
Besides, we apply some post-processing in inference to generate scores and singing voice for more tolerance (Appendix \ref{appendix:infer}).

\begin{table}[tbp]
    \centering
    \setlength{\tabcolsep}{2pt}
    \begin{tabular}{l|c|c|c|c}
    \hline
    \multirow{2}{*}{Models} & \multicolumn{2}{c|}{BLEU$\uparrow$} & \multicolumn{2}{c}{AS. $\uparrow$}\\
    \cline{2-5}
    & En$\rightarrow$Zh & Zh$\rightarrow$En & En$\rightarrow$Zh & Zh$\rightarrow$En \\
    \hline\hline
    GagaST & 11.87 & 5.67 & 0.701 & 0.468\\
    \hline
    \modelname-cls & 14.21 & 10.01 & 0.827 & 0.555\\
    ~~~ only bt  & 15.54 & 10.21 & 0.709 & 0.667\\
    ~~~ w/o bt & 13.73 & 8.26 & 0.704 & 0.490 \\
    \hline
    \modelname & 16.02* & \textbf{10.68} & \textbf{0.923} & \textbf{0.781} \\
    ~~~ only bt  & \textbf{16.27} & 10.26* & 0.880* & 0.718* \\ 
    ~~~ w/o bt & 14.12 & 7.86 & 0.845 & 0.710\\
    ~~~ w/o $\mathbf{e}_{align}$  & 15.16 & 9.24 & 0.852 & 0.703\\
    \hline
    \end{tabular}
    \caption{The sacreBLEU and Alignment Score on both translation directions. * means the second highest result within the row.}
    \label{tab:objective}
\end{table}

\subsection{Main Results}
We compare \modelname~with two baseline systems. 
One is the GagaST system~\cite{gagast}, which focuses on the tonal aspect of Chinese. 
The other one is a variation of our model. This variation uses a transformer-layer based classifier (\modelname-cls) instead of our alignment decoder to predict the number of aligned notes. The maximum number of aligned notes is 30, the same as allowed maximum $K(j)$ in alignment decoder. 
Besides, we show results from the human reference.
\subsubsection{Translation Evaluation}

We first report the human evaluation metrics (MOS-T) on both Chinese-to-English~(Zh$\rightarrow$En) and English-to-Chinese~(En$\rightarrow$Zh) song translation tasks in Table \ref{tab:subjective}. 
\modelname~generally gains improvements among all systems while the gap between different systems and settings is not obvious. 
It's partly because the lyrics translation by professionals is usually free translation rather than literal translation. 
A missing word in different slices can cause negative, neutral or even positive effect. 
Only obvious semantic deviations or grammatical mistakes lead to certain score decrease. 
As discussed in MOS-T, automatic metric BLEU may not be a good criterion to compare the machine translation and free translation for lyrics. 
But we still present the BLEU results in Table \ref{tab:objective}. 
We can see that our proposed system \modelname~significantly outperforms the recent baseline GagaST by a large margin on both translation directions. 
As to the variant model \modelname-cls, the \modelname~is still slightly better. 

%需要解释BLEU值 only 高，加了标注数据有所下降
\begin{figure*}[t]
    \centering
\subfigure[Source and Reference. Left: En$\rightarrow$Zh. Right: Zh$\rightarrow$En]{
    \includegraphics[width=0.55\textwidth,clip=true]{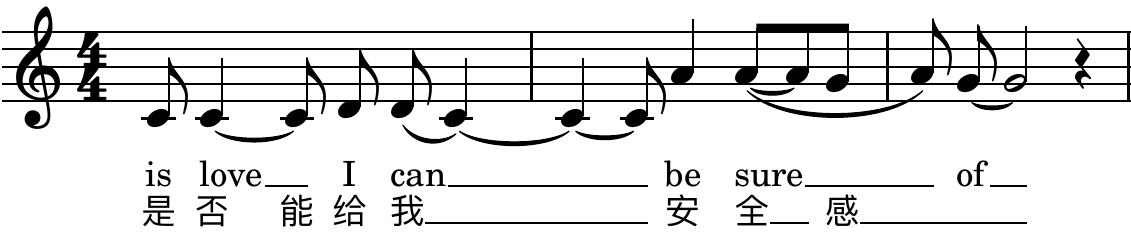}
    \includegraphics[width=0.43\textwidth,clip=true]{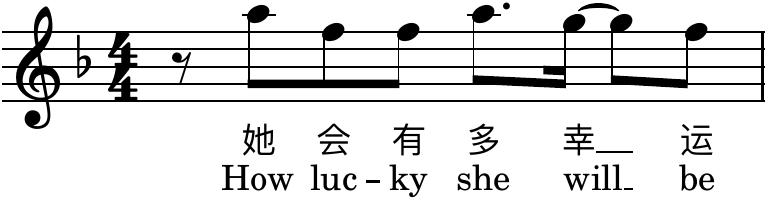}
}
\subfigure[GagaST]{
    \includegraphics[width=0.55\textwidth,clip=true]{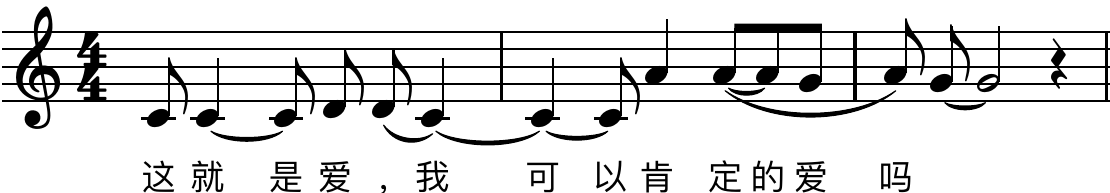}
    \includegraphics[width=0.44\textwidth,clip=true]{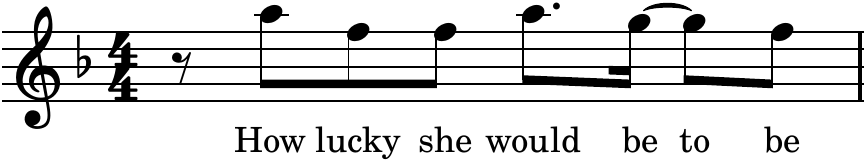}
}
\subfigure[\modelname-cls]{
    \includegraphics[width=0.55\textwidth,clip=true]{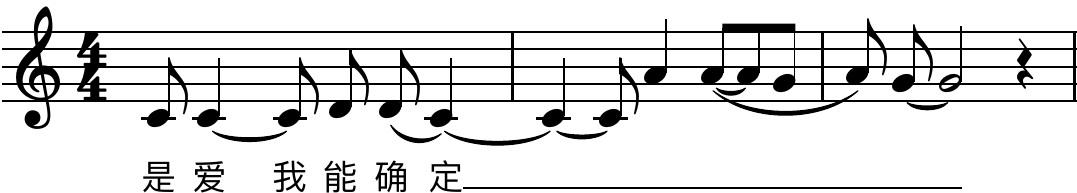}
    \includegraphics[width=0.44\textwidth,clip=true]{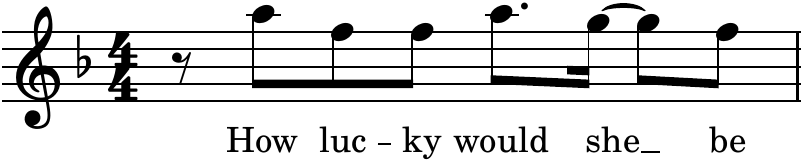}
}
\subfigure[\modelname]{
    \includegraphics[width=0.55\textwidth,clip=true]{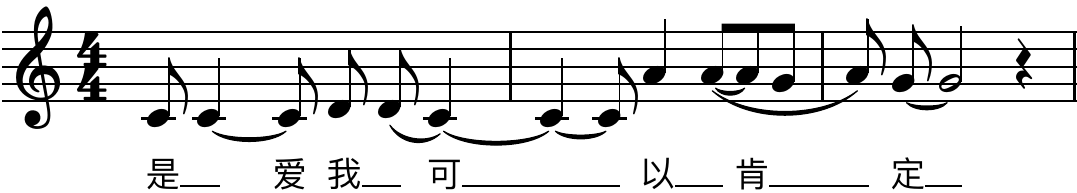}
    \includegraphics[width=0.44\textwidth,clip=true]{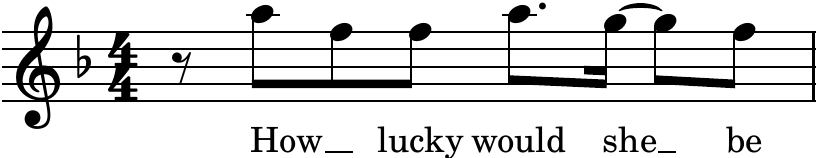}
}
\caption{Example scores of the source, reference and the translation for ``Is love I can be sure of'' in \textit{Will You Love Me Tomorrow} and ``t\={a} hu\.{i} y\v{o}u du\={o} x\.{i}ng y\.{u}n'' in \textit{Xi\v{a}o X\.{i}ng Y\.{u}n} from three systems.}
\label{fig:score_analysis}
\end{figure*}

\subsubsection{Lyrics-Melody Alignment Evaluation}

As for lyrics-melody alignment quality, we report the human evaluation metrics (MOS-S) on en-zh translation direction. 
In Table \ref{tab:subjective}, \modelname~considerably outperforms other systems, especially better than the GagaST with simple length control decoding. 
Notably, the variant version \modelname~cls performs worse than other systems, which indicates that more flexible alignments between lyrics and melody bring listening enjoyment to audience when it's reasonable enough. 
Otherwise, the flexibility may be counterproductive.  
We also evaluate the alignment quality by using the histograms of the number of aligned notes in Figure \ref{fig:align_hist}. 
In Table \ref{tab:objective}, we calculate the Alignment Score between the histograms of each system and the true histograms. 
The histograms show that the distribution of alignments generated by \modelname~resemble those of the true alignments while ``GagaST'' lacks variety by providing only one-on-one alignments between the lyrics and melody. 
In conclusion, both results demonstrate that the adaptive grouping method shows significant advantage over the length control or simple classifier in predicting reasonable alignment between the translated lyrics and melody. 

In Table \ref{tab:subjective}, MOS-Q mainly reflects the overall intelligibility, naturalness, singability and beauty of the song translation. 
Since the translation and alignment quality both contribute to the final result, the difference between methods seem less visible. 
But considering the 95\% confidence, we can conclude that the \modelname~still ranks best.

\subsection{Ablation Study and Analysis}

We first conduct ablation experiments to study the effects of back translation data with various settings. 
In Table \ref{tab:subjective}, we have the following findings for \modelname~and \modelname-cls. 
(1) Since the back-translation data is obviously larger than the real annotation data, there is almost no difference if only back-translation data is used for training. 
This enables the possibility of training our model in unsupervised way. 
(2) If only the limited supervised data is used, the performance apparently becomes worse. 
(3) \modelname~is consistently better than \modelname-cls in all ablation experiments. 
In addition, we verify the importance of the novel alignment embedding $\mathbf{e}_{align}$ by removing it from the note-pooling embedding layer and alignment decoder, and observe a non-negligible decrease on both BLEU and AS.

% As the sampling ratio of back-translation data goes down, the alignment quality increases while translation quality holds stable.
% \begin{table}[htbp]
%     \centering
%     \begin{tabular}{|c|c|c|}
%     \toprule
%     &  \multicolumn{2}{c|}{Co-translation Failure Rate}\\
%     \cmidrule{2-3}
%          & en-zh & zh-en \\
%     \midrule
%     GagaST & 0.0\% & 0.0\% \\
%     \midrule
%     TF+cls & 5.12\% & 4.23\% \\
%     \modelname & 6.78\% & 9.16\% \\
%     \bottomrule
%     \end{tabular}
%     \caption{Rates of translation failure of different systems.}
%     \label{tab:tarns_error}
% \end{table}

Some case studies in Figure \ref{fig:score_analysis} suggest that, when the tokens in lyrics fall into one-to-many alignments, GagaST usually provides inappropriate lyrics translation or even decodes non-vocal tokens such as comma to meet the length constraint. 
It will hurt both the translation quality and the singability of the translated lyrics. 
In contrast, the simple classifier following transformer layers is enough for flexible alignments. 
However, our evaluation results indicate our light weighted alignment decoder is capable of providing delicate alignments between tokens and notes.

%% file: appendix.tex
% \usepackage{times}
% \usepackage{latexsym}
% \usepackage[T1]{fontenc}
% \usepackage[utf8]{inputenc}

% % This is not strictly necessary, and may be commented out,
% % but it will improve the layout of the manuscript,
% % and will typically save some space.
% \usepackage{microtype}
% \usepackage{booktabs}
% \usepackage{multirow}
% \usepackage{multicol}
% \usepackage{subcaption}
% % \usepackage{subfigure}
% \usepackage[linesnumbered,ruled]{algorithm2e}
% \usepackage{mathtools,amssymb,mathrsfs}
% \usepackage{microtype}
% \usepackage{enumitem}
% \begin{document}

\appendix

\section{Pooling Matrix in the Note-pooling Embedding Layer}
\label{appendix:pool_mat}

We have note embedding $\mathbf{e}_{note} \in \mathbb{R}^{N \times d}$ ($d$ is the embedding dimension) and alignment matrix $\mathbf{M} \in \{0, 1\}^{L \times N}$. 
The non-overlapped mean-pooling can be calculated as follows.
\begin{align*}
\mathbf{W} &= \mathbf{M} / \text{sum}(\mathbf{M}, \text{dim}=-1, \text{keepdim}=\text{True}) \\
\mathbf{e}_{md} &= \mathbf{W} * \mathbf{e}_{note}
\end{align*}
where $/$ is element-wise division and $*$ is matrix multiplication. 
By leveraging \texttt{gather} and \texttt{scatter} operations, the non-overlapped mean-pooing can even be computed in batch.

\section{Analysis of Adaptive Grouping Loss}
\label{appendix:group_loss}

By the definition of the adaptive grouping loss, we only need to analyze the following term.
\begin{equation*}
    \left| K(j) - (1 - R(j)) - \Delta_j \right|
\end{equation*}

If $K(j) > \Delta_j$ in the forward pass, we have $K(j) - \Delta_j \geq 1$ because they are both positive integers. 
In order to encourage the loss to become smaller, $1 - R(j)=\sum_{k=1}^{K(j)-1}\alpha_j^k$ should become larger. 
In other words, the optimization will push $\sum_{k=1}^{K(j)-1}\alpha_j^k$ to be larger towards $K(j) - \Delta_j$. 
Note that the theoretical upper bound of $\sum_{k=1}^{K(j)-1}\alpha_j^k$ is $K(j)- 1$, which is larger or equal to $K(j)- \Delta_j$. 
Thus, this optimization is possible and it will meet the following condition during optimization.
\begin{equation*}
    \sum_{k=1}^{K(j)-1}\alpha_j^k \geq 1 - \epsilon .
\end{equation*}
By definition of $K(j)$, we have the following conclusion.
\begin{equation*}
    K(j)^{\text{new}} = \arg\min_{K}\left\{\sum_{k=1}^K \alpha_j^k \geq 1 - \epsilon\right\} \leq K(j) - 1
\end{equation*}

If $K(j) < \Delta_j$, a similar analysis can be derived. 
$\sum_{k=1}^{K(j)-1}\alpha_j^k\rightarrow0$ should be encouraged to purse a smaller loss. 
It implies if the $K(j)$-th halting probability doesn't satisfy the condition $\alpha_j^{K(j)} \geq 1-\epsilon$, the $K(j)^{\text{new}}$ will have an increasing trend.
However, if $\alpha_j^{K(j)} \geq 1 - \epsilon$, the optimization will be stuck. 
We may optimize $\left| K(j) - \left(1 - R(j) + \alpha_j^{K(j)}\right) - \Delta_j \right|$, \emph{i.e.}, $\left| K(j) - \sum_{k=1}^{K(j)} \alpha_j^{k} - \Delta_j \right|$. 
In practice, we found this is a rare case and the will completely disappear after several epochs. 
So we adopt the unified adaptive grouping loss.

If $K(j) = \Delta_j$, it means we can safely remove this term in the loss.

\section{Scheduler of Curriculum Learning}
\label{appendix:bt_cl}

The down sampling ratio of back translation data starts at 1.00 and decrease to 0.01 at the half of total training epochs.
The sampling ratio of annotation data starts at 20.00 for upsampling and decrease to 5.00 at the end of total training epochs.

\begin{figure}[htbp]
    \centering
    \includegraphics[width=0.5\textwidth]{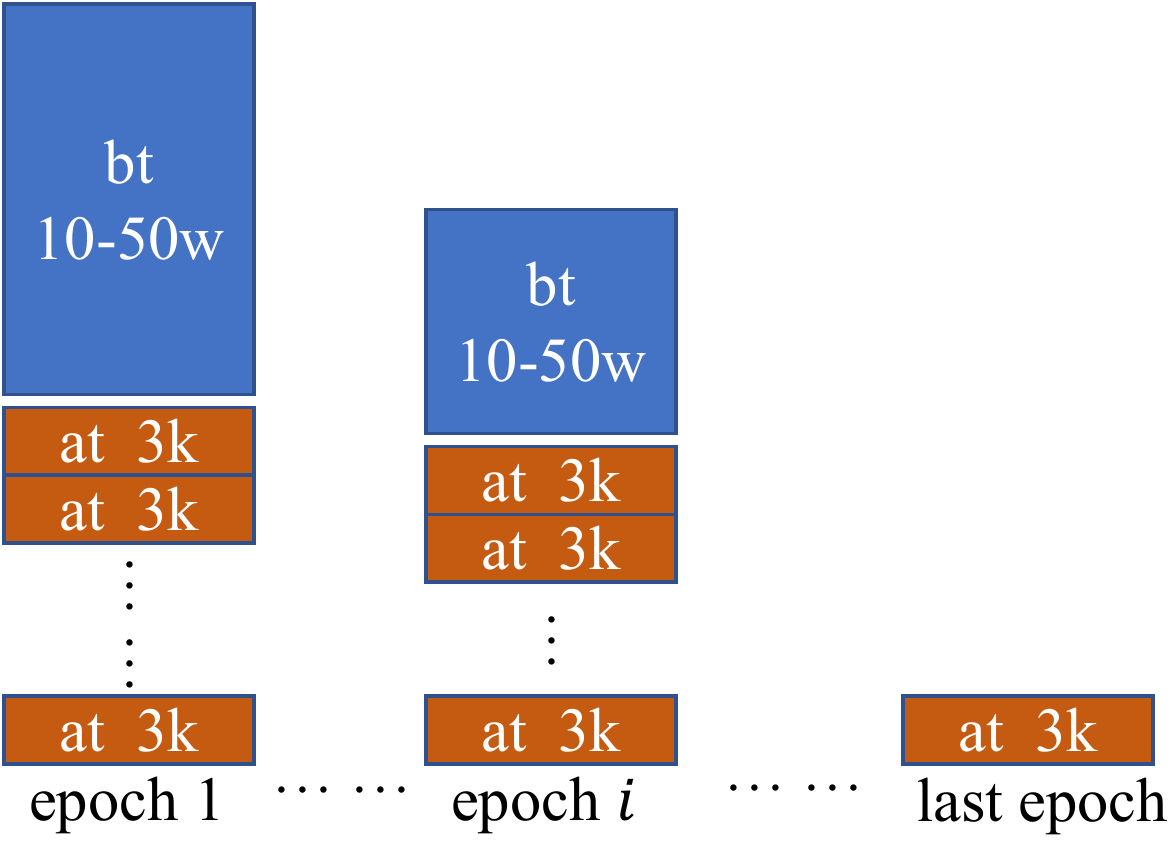}
    \caption{An illustration of how we use back translation data together with annotated data in co-translation training. ``bt'' represent data from back translation data augmentation and ``at'' represent data from annotation.}
    \label{fig:bt_curriculum}
\end{figure}

%\section{Limitations and Future Work}
%Future endeavors may lie in mining and utilizing song translation data from richer sources and more languages. 
%Besides, 
\section{Post-processing In Inference}
\label{appendix:infer}

In order to generate scores and singing voice in line with musical rules, we add some rule-based post-processing to the alignment predictions for more tolerance. 
For cases where total number of aligned notes is larger than the number of notes in the melody, we simply truncate the predicted number of aligned notes from the last token to the first or from the first token to the last. 
For cases of fewer number of predicted notes, we add the number of difference all to the last token.

\section{Data Annotation and Human Evaluation}
\label{appendix:data}

\begin{figure}[t]
    \centering
    \includegraphics[width=0.49\textwidth]{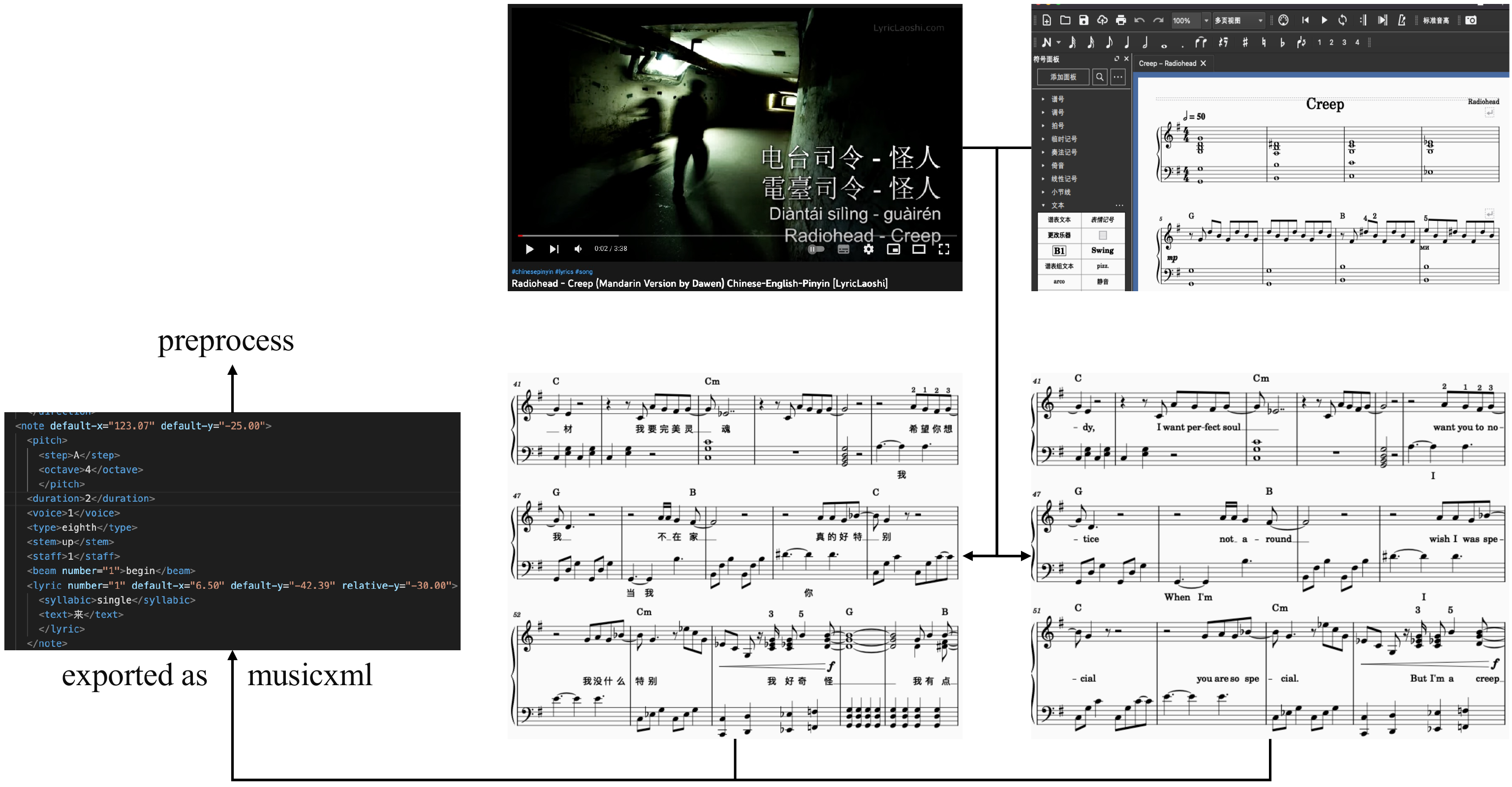}
    \caption{An overview of our propdata annosed otation pipeline.}
    \label{fig:da_pipeline}
\end{figure}

Annotators are students who major in music, vocal singing or relevant specialty. 
They all speak bilingual languages with Chinese and English, so they are also qualified for translation quality evaluation. 
For data annotation and human evaluation, each person gets reasonably paid according to the individual workload.
The annotation guidance and evaluation guidance can be found in supplement materials. 
Figure \ref{fig:eval_page} is an example of visual front-end interface for human evaluation. The pipeline for data annotation is shown in Figure \ref{fig:da_pipeline}

\begin{figure}[t]
    \centering
    \includegraphics[width=0.49\textwidth]{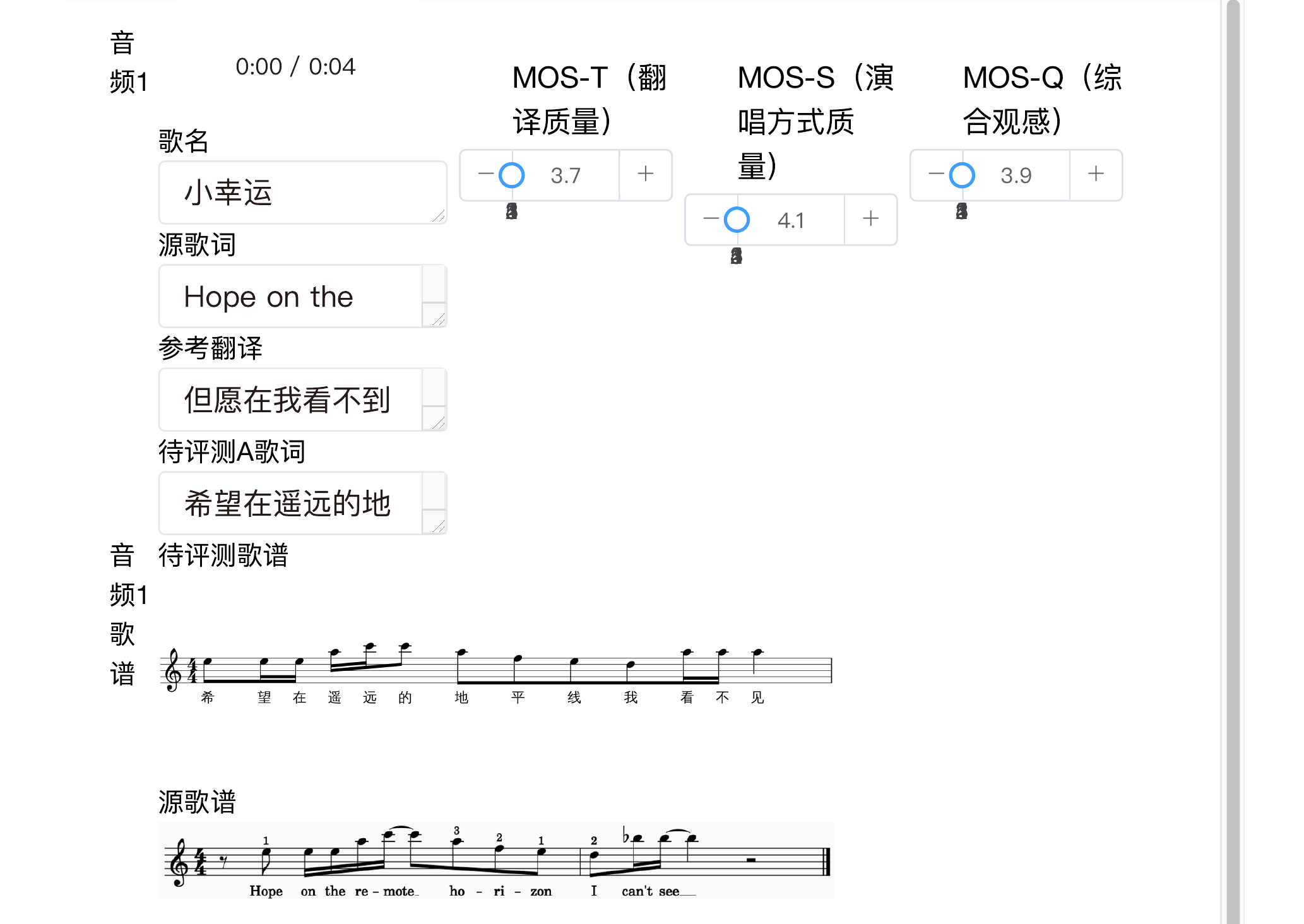}
    \caption{An example of evaluation front-end interface for human evaluation.}
    \label{fig:eval_page}
\end{figure}

\section{Chinese-to-English Song Translation Evaluation}
\label{appendix:zh-en}

Here we show the MOS-S and MOS-Q for Chinese-to-English song translation for reference. 
Lack of open data set for English singing voice synthesis caused the bad quality of synthesized English singing voice in inference. 
So we have to use the Chinese SVS system to synthesize the English songs. 
According to the feedback from annotators, this gap influence their feeling about translation results to some extent. As is shown in Table \ref{tab:zh_en_subjective}, results significantly drop compared to those of En$\rightarrow$Zh.
So we leave this to appendix part for reference.

\begin{table}[htbp]
    \centering
    \begin{tabular}{l|c|c}
    \hline
    & MOS-S & MOS-Q \\
    \hline
    & Zh$\rightarrow$En & Zh$\rightarrow$En\\
    \hline
    Human Ref. & 4.05 $\pm$ 0.08 & 4.06 $\pm$ 0.07\\
    \hline
    GagaST & 3.04 $\pm$ 0.02 & 3.08 $\pm$ 0.03\\
    \hline
    \modelname-cls  & 3.03 $\pm$ 0.02 & 3.10 $\pm$ 0.01\\
    ~~~ only bt & 3.11 $\pm$ 0.04& 3.15 $\pm$ 0.03\\
    ~~~ w/o bt & 3.27 $\pm$ 0.04 & 3.20 $\pm$ 0.02\\
    \hline
    \modelname  & 3.39 $\pm$ 0.03& 3.61 $\pm$ 0.05\\
    ~~~ only bt & 3.39 $\pm$ 0.05 & 3.60 $\pm$ 0.02\\
    ~~~ w/o bt  & 3.29 $\pm$ 0.05 & 3.32 $\pm$ 0.02\\
    % \midrule
    % \modelname~w/o bt  & & & & & & \\
    % \modelname~only bt & & & & & & \\
    % \modelname~+ bt  & & & & & & \\
    \hline
    \end{tabular}
    \caption{The  Mean  Opinion  Score singability~(MOS-S) and overall quality~(MOS-Q) for Zh$\rightarrow$En samples with 95\% confidence intervals.}
    %带*的结果仅供参考
    \label{tab:zh_en_subjective}
\end{table}